\documentclass[letterpaper, 10 pt, journal, twoside]{IEEEtran}

\usepackage{cite}

\ifCLASSINFOpdf
  \usepackage[pdftex]{graphicx}
\else
\fi

\usepackage{amsmath}
\usepackage{bm}
\usepackage{algorithm}
\usepackage{algorithmic}
\usepackage{array}
\usepackage{xcolor}
\usepackage{url}
\usepackage{enumitem}
\usepackage{booktabs}
\usepackage{multirow}
\usepackage{multicol}

\usepackage{kotex}
\usepackage{float}
\usepackage{placeins}

\usepackage{cuted}
\usepackage{capt-of}

\usepackage{graphicx}
\usepackage{array}
\newcommand{\ExperimentAssetPath}{images}
\newlength{\ExpLabelWidth}
\newlength{\ExpGap}
\newlength{\ExpFourWidth}
\newlength{\ExpThreeWidth}
\newcommand{\ExpPhoto}[3]{%
  \begin{minipage}[c]{#1}%
    \includegraphics[width=\linewidth,trim={#2},clip]{\ExperimentAssetPath/#3}%
  \end{minipage}%
}

\begin{document}
\raggedbottom

\title{Now You Feel It, Now You See Me:\\
Digital-Twin-based Teleoperation Interface \\ for Dexterous Manipulation}

\author{Youngchan Shim\textsuperscript{1,*}, Kyutae Lee\textsuperscript{1,*}, JooYun Kim\textsuperscript{1}, Jaeseong Hwang\textsuperscript{1}, Harim Ji\textsuperscript{2}, Yongseok Lee\textsuperscript{1,\(\dagger\)}
\thanks{\textsuperscript{1}Youngchan Shim, Kyutae Lee, JooYun Kim, Jaeseong Hwang, and Yongseok Lee are with the Department of Robotics and Mechatronics Engineering, DGIST, Daegu, Korea.}
\thanks{\textsuperscript{2}Harim Ji is with the Department of Mechanical Engineering, Seoul National University, Seoul, Korea.}
\thanks{\textsuperscript{*}These authors contributed equally. \textsuperscript{\(\dagger\)}Corresponding author: Yongseok Lee (email: yslee@dgist.ac.kr).}}
\markboth{}{}
\maketitle
\begin{strip}
\centering
\includegraphics[width=\textwidth]{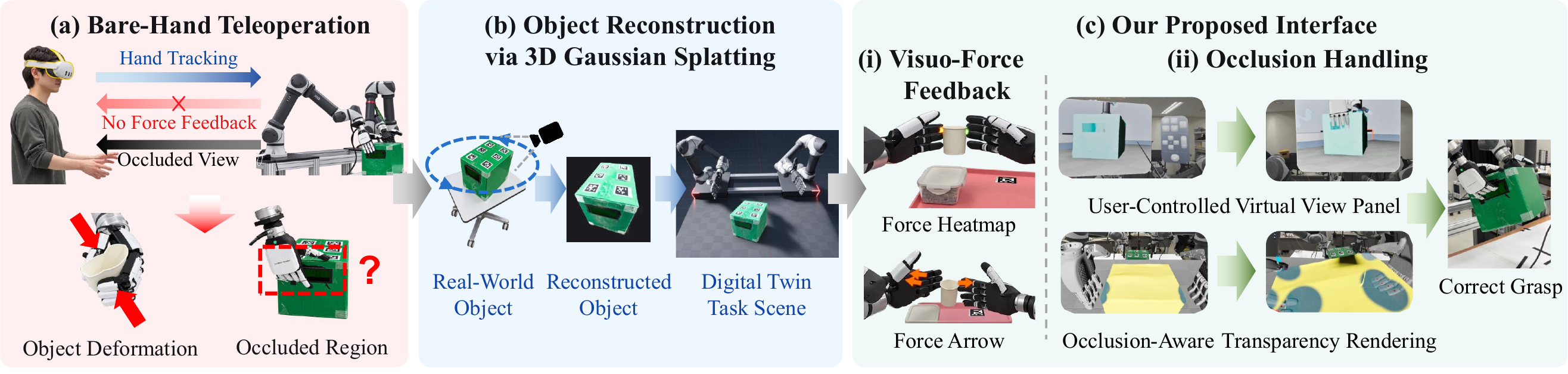}
\captionof{figure}{Overall configuration of the proposed Digital-Twin-based Augmented Reality (AR) teleoperation interface for bimanual multi-fingered robotic hands.}
\label{fig:overview}
\end{strip}

\begin{abstract}

Teleoperation is becoming increasingly important for collecting high-quality demonstrations to teach robots dexterous manipulation skills. For dexterous manipulation, bare-hand tracking provides a practical way to control robotic hands and demonstrate coordinated finger movements without gloves or exoskeletons. However, this type of teleoperation faces two key feedback limitations: a lack of force feedback and visual feedback of occluded region. The absence of force feedback hinders precise and safe manipulation, as operators must infer contact force visually rather than feel them directly. Occlusion by objects or other robot parts impairs the assessment of hand positions, approach distances, and pre-grasp configurations suited to the object's shape. To address these limitations, we propose a digital-twin-based augmented reality (AR) teleoperation interface that integrates bare-hand tracking, bimanual robotic hands, and virtual representations of the robot and task objects. The interface renders tactile measurements on robot hand meshes as visuo-force feedback. Furthermore, it offers a user-controlled virtual view panel or occlusion-aware transparency rendering to provide occlusion-mitigating visual feedback. We evaluated the interface through user studies involving Task 1 and Task 2, examining how force visualization and visual assistance support demonstration collection in tasks requiring careful force regulation and manipulation under occlusion.
\end{abstract}
\IEEEpeerreviewmaketitle
\section{Introduction}

Teleoperation has become a primary means of collecting high-fidelity demonstration data for robot learning by recording robot actions and observations during human-guided manipulation. Specifically, for dexterous robot hand manipulation, human demonstrations provide high-dimensional joint trajectories during interactions with diverse objects for robot learning. Vision-based hand tracking and motion retargeting enable operators to control robotic hands through bare-hand movements across different robot configurations~\cite{qin2023anyteleop,iyer2024open}. By eliminating the need for gloves or exoskeletons for motion input, bare-hand interfaces provide a practical means of demonstrating coordinated finger movements.

Despite these advantages, bare-hand teleoperation with typical visual feedback (i.e., bare-eyes or camera) provides limited information about hand--object interactions. In a headset-based setup, operators control the robot through tracked hand movements while observing the physical workspace through passthrough. Without force feedback, they must infer finger contact and contact forces from visual cues, hindering efforts to prevent slippage or excessive grasp forces. Moreover, occlusion by objects or other robot parts obscures finger placement and approach distances, hindering grasp formation. Collectively, these limitations impede the monitoring and correction of manipulation errors during demonstration collection.

Prior work has addressed the lack of force information through wearable haptic devices or visual feedback. Vibrotactile, cutaneous or exoskeleton haptic devices on Glove can convey contact forces, but require hand-worn hardware that can restrict finger motion and complicate deployment across operators~\cite{DOGlove,ding2025bunny, wchi}. As a visual alternative, TactAR~\cite{xue2025reactive} renders a tactile sensor's deformation field in AR and reports improved demonstration quality relative to teleoperation without this feedback. However, how to represent per-finger force information for dexterous hand teleoperation remains underexplored, particularly the relative benefits of force heatmap and force arrow representations.

To address occlusion, previous studies position or move physical cameras to obtain alternative views~\cite{rakita2019remote} or render the robot or selected links transparent to reveal hidden regions~\cite{taylor2020diminished,li2026xr}. However, these studies neither directly compare the two approaches in dexterous hand teleoperation nor establish their interaction with force visualization.

We propose digital-twin-based AR interfaces combining visuo-force feedback with occlusion-mitigating visual feedback for bare-hand bimanual dexterous teleoperation. Digital-twins of the robot and task objects are updated according to robot states and tracked object poses to visualize contact forces and occluded spatial relationships. We design two force representations: (1) a force heatmap encoding tactile signal magnitude through pad-level color and (2) force arrows conveying direction and magnitude through fingertip-anchored arrows (Fig.~\ref{fig:force_viz}). For occlusion handling, we introduce (1) a user-controlled virtual view panel, which displays an operator-controlled virtual camera view alongside passthrough, and (2) occlusion-aware transparency rendering, which reveals robot fingers through semi-transparent objects in the digital-twin layout (Fig.~\ref{fig:experimental-interfaces}). These interfaces support force regulation and grasp formation without hand-worn feedback devices.

We evaluated the interfaces through a user study assessing task performance and perceived workload. The study compared force heatmap, force arrow, and no visuo-force feedback in Task 1, which requires careful force regulation with unobstructed contact regions. In Task 2, we combined the force visualization evaluated in the Task 1 user study with the user-controlled virtual view panel and occlusion-aware transparency rendering. We analyzed and evaluated its interaction with each occlusion-assistance technique during bimanual lifting with occluded grasp openings. In Task 1, both force visualizations significantly reduce perceived workload relative to no visuo-force feedback. The cup deformation rate is 13.3\% with force arrow feedback, compared with 44.4\% without visuo-force feedback. In Task 2, occlusion-aware transparency rendering achieves the highest observed success rate (82.1\%) when combined with force arrow feedback.

The contributions of this work are as follows:
\begin{enumerate}
    \item We propose digital-twin-based AR interfaces that combine per-finger visuo-force feedback with a user-controlled virtual view panel or occlusion-aware transparency rendering for bare-hand bimanual dexterous teleoperation.
    \item We compare force heatmap feedback, force arrow feedback, and no visuo-force feedback during manipulation requiring precise grasp force regulation.
    \item We evaluate the two occlusion-assistance techniques for supporting manipulation under occlusion and evaluate them with and without visuo-force feedback.
\end{enumerate}

\section{Related Works}

\subsection{Robot Hand Teleoperation}

AnyTeleop~\cite{qin2023anyteleop} supports vision-based arm--hand teleoperation across different robot arm, hand, and camera configurations, emphasizing adaptability across hardware setups. Open Teach~\cite{iyer2024open} combines HMD hand tracking with close-up visual feedback for multi-fingered and bimanual control. Open-TeleVision~\cite{cheng2024open} couples arm and hand motion transfer with active stereoscopic feedback, integrating motion input with immersive workspace observation. Alongside these vision-based approaches, wearable gloves provide an alternative means of tracking hand motion for teleoperation~\cite{vist,manus}.

While these approaches support motion input and task observation, hand tracking alone does not convey robot contact forces, and scene views may leave hand--object interactions occluded. Our bare-hand AR interfaces therefore combine visuo-force feedback with occlusion-mitigating visual feedback to support perception of forces and hand--object spatial relationships during bimanual manipulation.

\subsection{Tactile Feedback for Human Operators}

Wearable haptic devices convey contact information through vibration~\cite{ding2025bunny} or combine hand tracking with force feedback for robotic manipulation~\cite{DOGlove} and virtual manipulation~\cite{wchi}. Visual feedback studies compare color-based force representations in robot-assisted surgery~\cite{aviles2018sensory} and display forces as arrows at the end effector~\cite{hartwig2023visualization}. TactAR~\cite{xue2025reactive} presents tactile and force information as an AR deformation field registered to the robot end-effector.

Wearable approaches require hand-worn hardware, while the cited visual approaches examine surgical interaction, end-effector force display, or parallel-jaw grasping rather than a common dexterous hand setting. Their evaluations therefore do not directly establish how force heatmap and force arrow representations compare in bimanual dexterous teleoperation. We compare these representations with a baseline without force visualization on a task requiring precise grasp force regulation.

\subsection{Occlusion-Mitigating Visual Feedback}

Prior systems employ an assistant robot arm to adjust camera viewpoints~\cite{rakita2019remote} or a drone to provide task-relevant views~\cite{droneviewpoint}. VICARIOS~\cite{naceri2021vicarios} enables viewpoint changes within a virtual environment. Other approaches reveal occluded regions by rendering the robot transparent in diminished reality~\cite{taylor2020diminished} or rendering selected digital-twin robot links semi-transparent during task demonstration~\cite{li2026xr}.

Although these studies explore viewpoint adjustment and transparency, they do not directly compare the two approaches within a common dexterous hand teleoperation setting. These studies also do not establish the collaboration effect between force feedback and occlusion-mitigating feedback. We therefore compare a user-controlled virtual view panel in a passthrough interface with occlusion-aware transparency rendering in a digital-twin interface, alongside a baseline without occlusion handling. Each interface condition is evaluated with and without visuo-force feedback during bimanual manipulation under occlusion.

\section{Interface Design}

\subsection{Object Reconstruction for the Digital-Twin}
We adopt 3D Gaussian Splatting (3DGS)~\cite{kerbl3Dgaussians} to reconstruct the physical task objects for the digital-twin environment because it supports photorealistic novel-view rendering in real time. HIL-GS~\cite{lee2025human} highlights the value of interactive 3D visualization for providing spatial context beyond the limited perspectives of live camera streams in robotic teleoperation. In our interface, this capability motivates the adoption of 3DGS to preserve object appearance while allowing operators to inspect hand--object spatial relationships from viewpoints beyond their direct view. We construct each object's representation from images captured around it, enabling virtual viewpoint changes without placing or moving a physical camera at the selected viewing location.

To obtain a surface representation alongside Gaussian splatting, we deploy the reconstruction and refinement pipeline of RORA~\cite{lee2026rora}, a refined version of GS2Mesh~\cite{wolf2024gsmesh} which can reconstruct more refined meshes than GS2Mesh. The pipeline builds on GS2Mesh for initial reconstruction and refines the resulting mesh by removing isolated fragments, reconstructing the surface, remeshing, and filling holes to obtain watertight geometry. These steps address missing surfaces and reconstruction artifacts in the initial mesh. The resulting Gaussian splatting and mesh assets are integrated into the virtual task scene.

During teleoperation, the virtual objects are positioned according to the tracked object state so that they can be displayed together with the virtual robot. The reconstructed scene supports the user-controlled virtual view panel in the passthrough interface and occlusion-aware transparency rendering in the digital-twin interface, as described in the following subsections.

\subsection{Visuo-Force Feedback}
\begin{figure}[!tbp]
\centering
\footnotesize
\setlength{\tabcolsep}{0pt}
\setlength{\fboxsep}{0pt}
\begin{tabular}{@{}c@{\hspace{0.04\linewidth}}c@{}}
\textbf{XHAND 1 (Left Hand)} & \textbf{Inspire Hand (Right Hand)} \\[4pt]
\multicolumn{2}{c}{\textbf{Force Heatmap}} \\[3pt]
\colorbox{black}{\includegraphics[width=3.2cm,height=3.2cm,keepaspectratio,trim={0.375bp 0bp 0.375bp 0bp},clip]{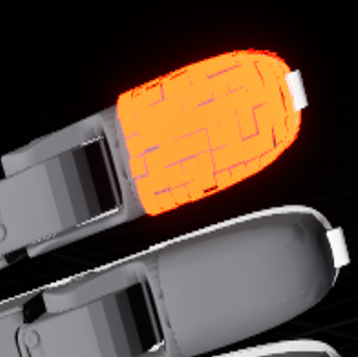}} &
\colorbox{black}{\includegraphics[width=3.2cm,height=3.2cm,keepaspectratio,trim={1.125bp 0bp 1.125bp 0bp},clip]{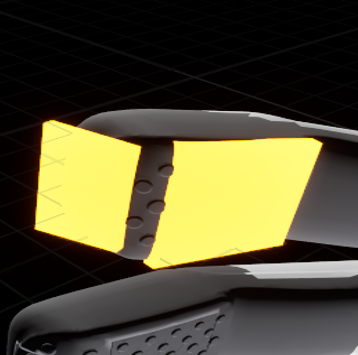}} \\[6pt]
\multicolumn{2}{c}{\textbf{Force Arrow}} \\[3pt]
\colorbox{black}{\includegraphics[width=3.2cm,height=3.2cm,keepaspectratio,trim={0.75bp 0bp 0.75bp 0bp},clip]{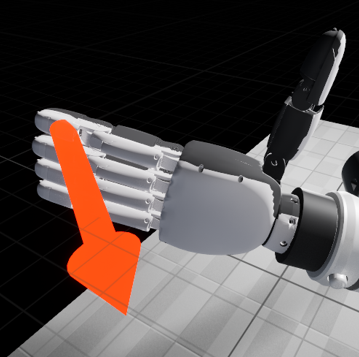}} &
\colorbox{black}{\includegraphics[width=3.2cm,height=3.2cm,keepaspectratio,trim={3.375bp 0bp 3.375bp 0bp},clip]{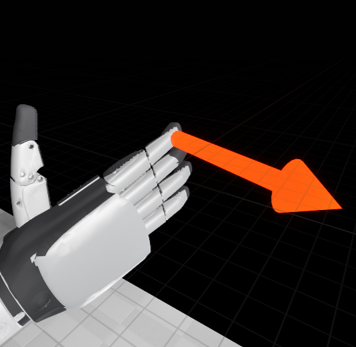}}
\end{tabular}
\caption{Visuo-force feedback representations for the left XHAND 1 (left column) and right Inspire Hand (right column). The top row presents force heatmap feedback, which encodes tactile signal magnitude through pad-level color. The bottom row presents force arrow feedback, which conveys force direction and magnitude through fingertip-anchored arrows.}
\label{fig:force_viz}
\end{figure}

We design the visuo-force feedback interface to be independent of a specific robot platform, with sensor-to-visual mappings that can be adapted to different robotic hands. The interface employs fingertip tactile measurements to provide contact information through two visual representations: force heatmap and force arrow. Figure~\ref{fig:force_viz} presents both representations on the left and right robotic hands.

\subsubsection{Force Heatmap}
Drawing on force-to-color mappings employed in robot-assisted surgery~\cite{aviles2018sensory}, we apply color-based force feedback to individual fingers of a multi-fingered robotic hand. Each fingertip sensor pad is uniformly colored according to its tactile signal magnitude to indicate contact and its intensity.

\subsubsection{Force Arrow}
Following prior force visualization methods~\cite{hartwig2023visualization,chan2018multimodal}, we apply arrow-based force feedback to individual fingers of a multi-fingered robotic hand. Each arrow is displayed at the corresponding fingertip to convey force direction and magnitude. arrows are rendered with emissive materials.

\subsection{Occlusion Handling}
We design two approaches to provide occlusion-mitigating visual feedback: (1) a user-controlled virtual view panel in the passthrough layout and (2) occlusion-aware transparency rendering in the digital-twin layout. Both approaches utilize a virtual scene implemented in Unreal Engine 5.3 and displayed through the operator's Meta Quest 3 headset.
\subsubsection{User-Controlled Virtual View Panel}
Inspired by camera-viewpoint adjustment in telemanipulation~\cite{rakita2018autonomous,rakita2019remote}, the user-controlled virtual view panel allows the operator to control the pose of the virtual camera within the reconstructed digital-twin task scene. The display panel renders the reconstructed virtual scene, allowing operators to select a virtual viewpoint without placing or moving a physical camera at the selected location. The rendered view is displayed alongside the passthrough view of the physical workspace.

The virtual robot is initially aligned with the physical robot in passthrough, after which the virtual scene is updated according to robot states and the tracked object pose. During the approach to the task object, the interface monitors the distance between the virtual robot and the virtual object. When this distance falls at or below a predefined threshold, teleoperation is automatically paused. Visual references of the operator's tracked hands are retained at their poses at the moment of pausing, and the display panel and camera-control buttons automatically appear in the operator's field of view (Fig.~\ref{fig:experimental-interfaces}).

The operator translates and rotates the virtual camera through the buttons and selects a viewpoint for the subsequent manipulation. Pressing the confirmation button hides the buttons and places the panel at the center of the operator's field of view. Teleoperation remains paused until the operator aligns their tracked hands with the retained hand references, at which point control resumes immediately. The panel then provides the selected virtual view during task execution. The interface does not support reselecting the viewpoint after confirmation
.
\subsubsection{Occlusion-Aware Transparency Rendering}

Occlusion by the task object obscures finger placement and hand--object contact regions, hindering grasp formation. We therefore present the virtual robot hands and task object in front of the operator, preserving their relative positions and orientations to support egocentric observation of hand--object spatial relationships. The operator's hand movements are mapped to relative changes in position and orientation within this configuration. The virtual scene is updated according to wrist poses received from the robot and the tracked object pose.

When the fingers are sufficiently far from the object, the object is rendered opaque to convey its shape and preserve depth-ordering cues. As a fingertip approaches, the interface locally reduces object opacity around the viewing ray from the operator to that fingertip, with stronger transparency at shorter distances. This reveals otherwise occluded fingers and regions near contact while retaining the surrounding object surface.

The transparent regions follow changes in fingertip positions and the operator's viewpoint, with multiple regions applied when several fingers approach the object. A semi-transparent object silhouette is displayed together with the local transparency to retain a visual reference for the object's overall shape and spatial extent. Both effects are removed as the fingers move away.

To provide additional spatial cues, red markers indicate the object surface points closest to reference points on the virtual fingertips and palms. Identifying contact locations enables the operator to assess which parts of the object each finger supports and adjust hand position and orientation during grasp formation. The markers support this assessment by visualizing the correspondence between occluded hand regions and the object surface. They are updated as the hands move and remain visible through the object and hands. 

The transparency rendering updates automatically during manipulation without requiring viewpoint selection or confirmation. When visuo-force feedback is enabled, force arrows display tactile measurements at the corresponding virtual fingers. The operator can therefore observe finger placement relative to the object together with the force magnitude measured at each finger. Local transparency, the object silhouette, and the surface markers remain available when visuo-force feedback is disabled.

\subsection{Hardware Setup}
\label{sec:hardware_setup}
\begin{figure}[!tbp]
\centering
\includegraphics[width=\linewidth]{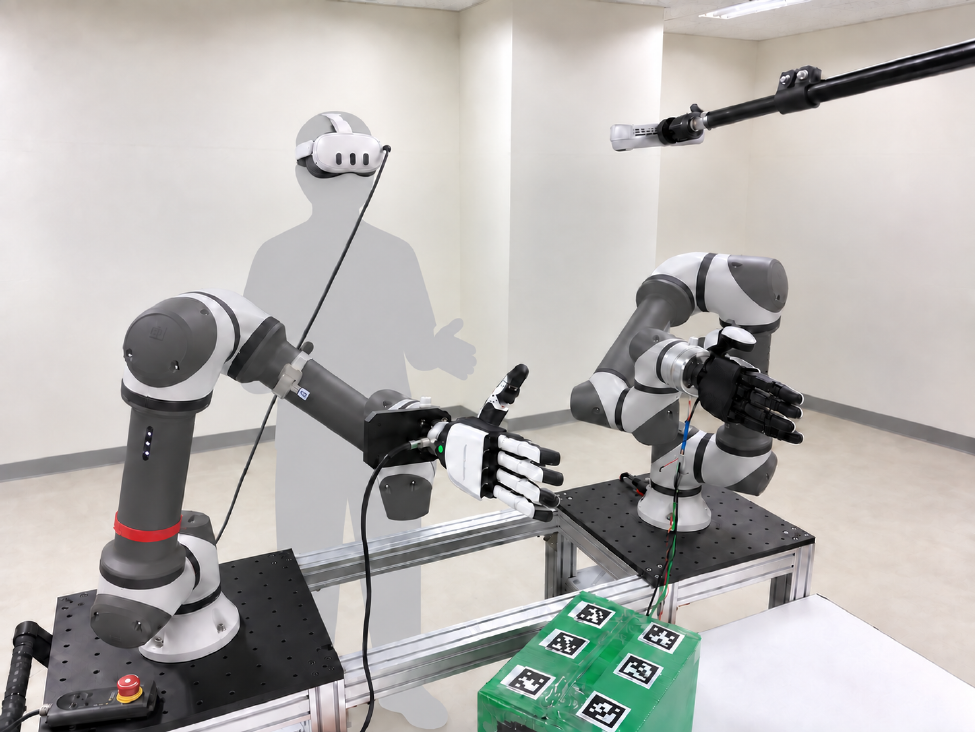}

\caption{Hardware setup. Two RB5-850E arms were equipped with an XHAND 1 (left) and an Inspire Hand RH56DFTP (right). The operator wore a Meta Quest 3 and controlled the hands with bare-hand tracking.}
\label{fig:hardware}
\end{figure}

Our bimanual setup, illustrated in Fig.~\ref{fig:hardware}, comprised two RB5-850E arms equipped with an XHAND 1 for the left hand and an Inspire Hand RH56DFTP for the right hand. The operator controlled the hands through bare-hand tracking with a Meta Quest 3 headset.

To accommodate the different tactile sensor layouts, the feedback utilized the left hand's tip pad measurements and the right hand's tip and nail pad measurements. For force heatmap feedback, the right hand's two pads were colored independently. For force arrow feedback, their signals were mapped to a single arrow at the corresponding fingertip, whereas the left hand's arrow was derived from its tip pad signal.

An ArUco marker attached to the physical task object was employed to track its pose and update the corresponding virtual object's position and orientation, allowing the virtual representation to follow the physical object's motion.

\section{Experiment Design}

\subsection{Participants}

A total of 15 participants (13 male, 2 female) were recruited from a university. Their mean age was 25.0 years (SD = 2.0, range: 21–28). All 15 participants took part in Task 1. Of these, 14 also participated in Task 2; one participant was unable to attend Task 2. Regarding prior experience, 2 participants (13.3) had previously operated a teleoperated robot, and 2 participants (13.3) had prior experience with a VR head-mounted display.

\subsection{Tasks}

\begin{figure*}[t]
  \centering
\begingroup
\fontsize{9}{10.5}\selectfont
\raggedright
\setlength{\tabcolsep}{0pt}
\renewcommand{\arraystretch}{1}
\setlength{\ExpLabelWidth}{0.15091016\linewidth}
\setlength{\ExpGap}{0.00783945\linewidth}
\setlength{\ExpFourWidth}{\dimexpr(\linewidth-\ExpLabelWidth-3\ExpGap)/4\relax}
\setlength{\ExpThreeWidth}{\dimexpr(\linewidth-\ExpLabelWidth-2\ExpGap)/3\relax}
\noindent{\fontsize{10}{12}\selectfont\bfseries (a) Task 1: Paper Cup Relocation}\par
\vspace{7pt}
\noindent\hspace*{\ExpLabelWidth}%
\begin{tabular}{@{}c@{\hspace{\ExpGap}}c@{\hspace{\ExpGap}}c@{\hspace{\ExpGap}}c@{}}
Reference & No Visuo-Force Feedback & Force Heatmap & Force Arrow \\[4pt]
\ExpPhoto{\ExpFourWidth}{154.25329bp 7.78702bp 114.25329bp 186.94702bp}{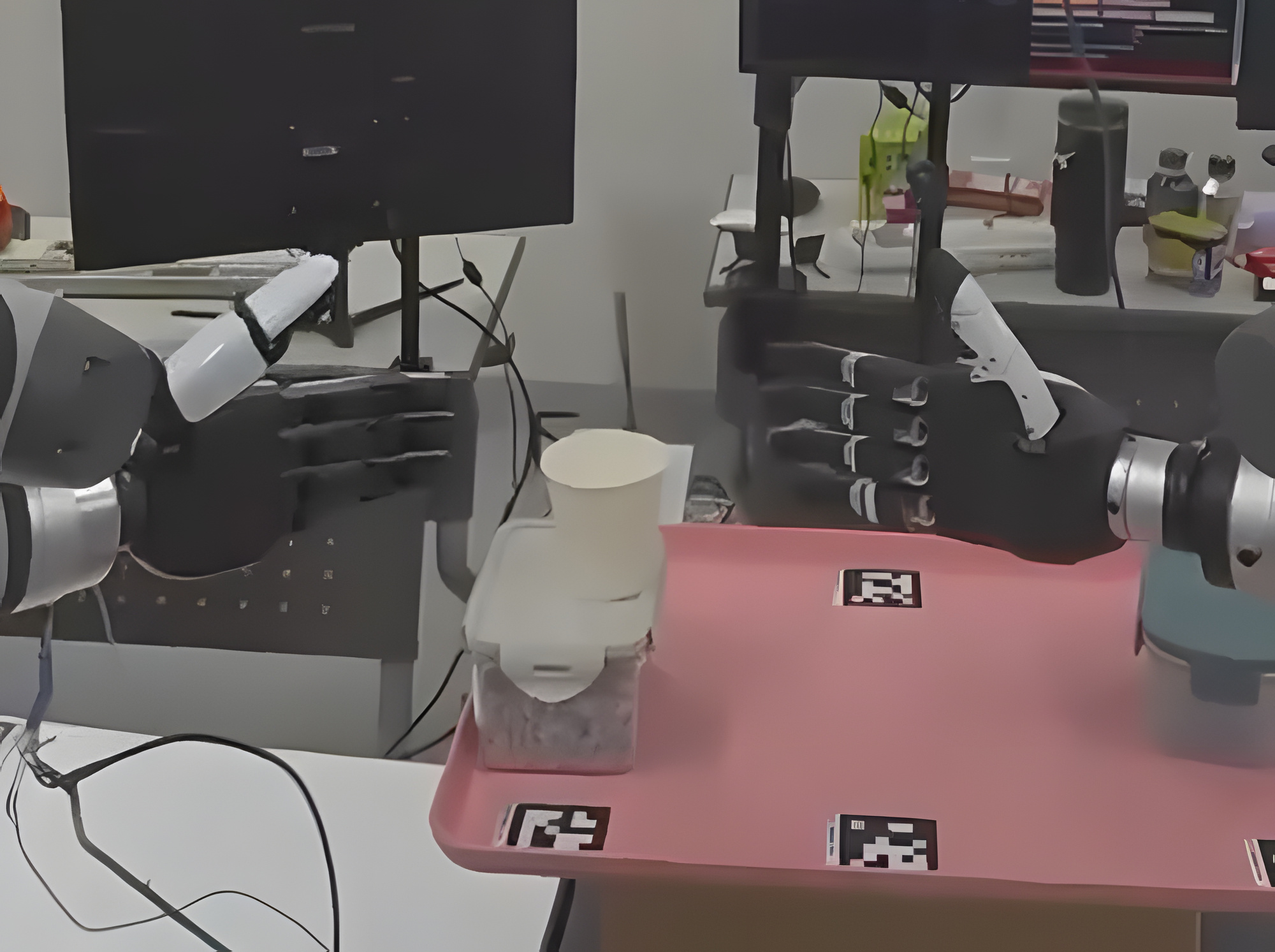} &
\ExpPhoto{\ExpFourWidth}{0.00000bp 84.54000bp 0.00000bp 56.46000bp}{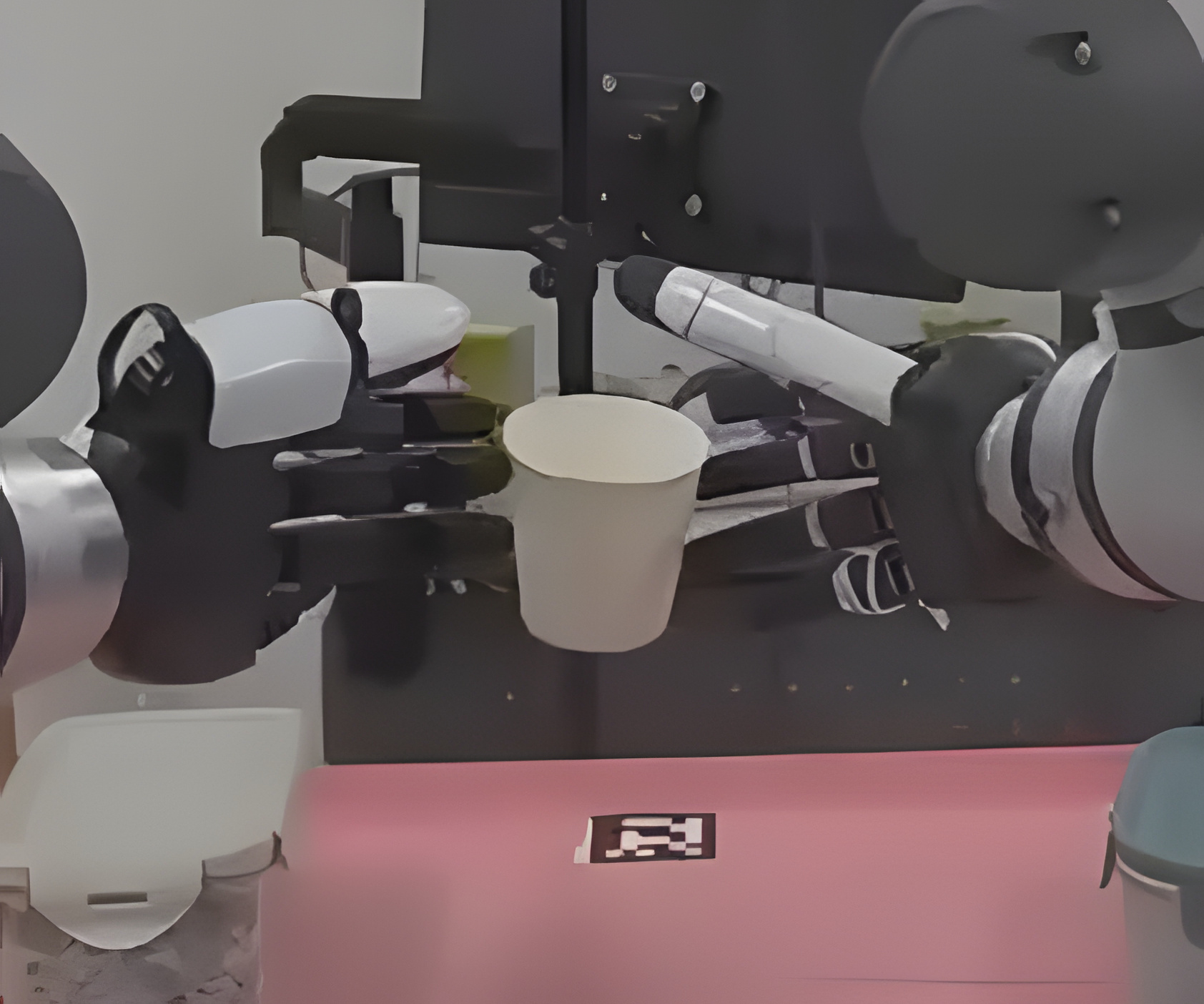} &
\ExpPhoto{\ExpFourWidth}{121.36811bp 0.00000bp 41.36811bp 0.00000bp}{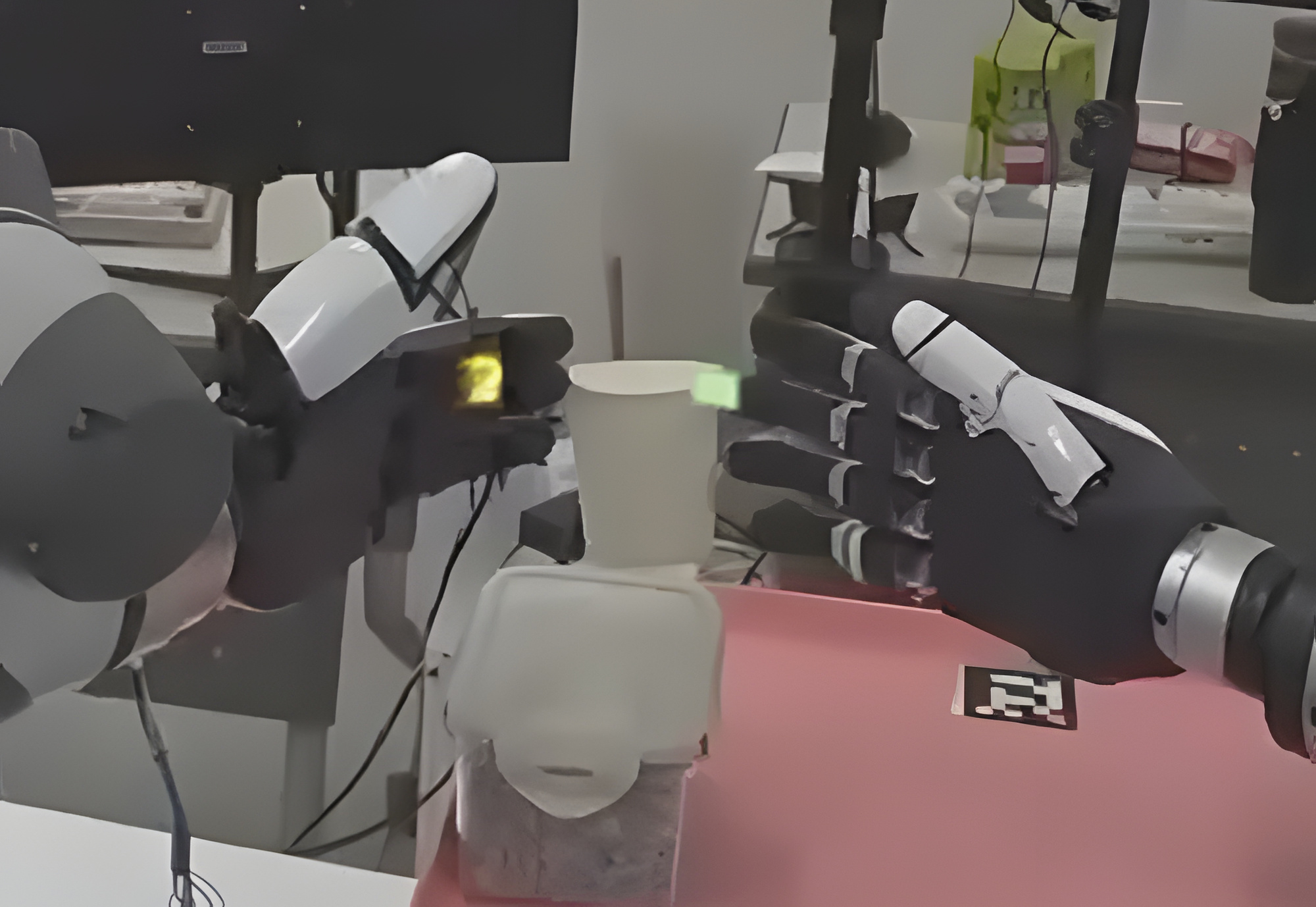} &
\ExpPhoto{\ExpFourWidth}{0.00000bp 28.52304bp 0.00000bp 60.30304bp}{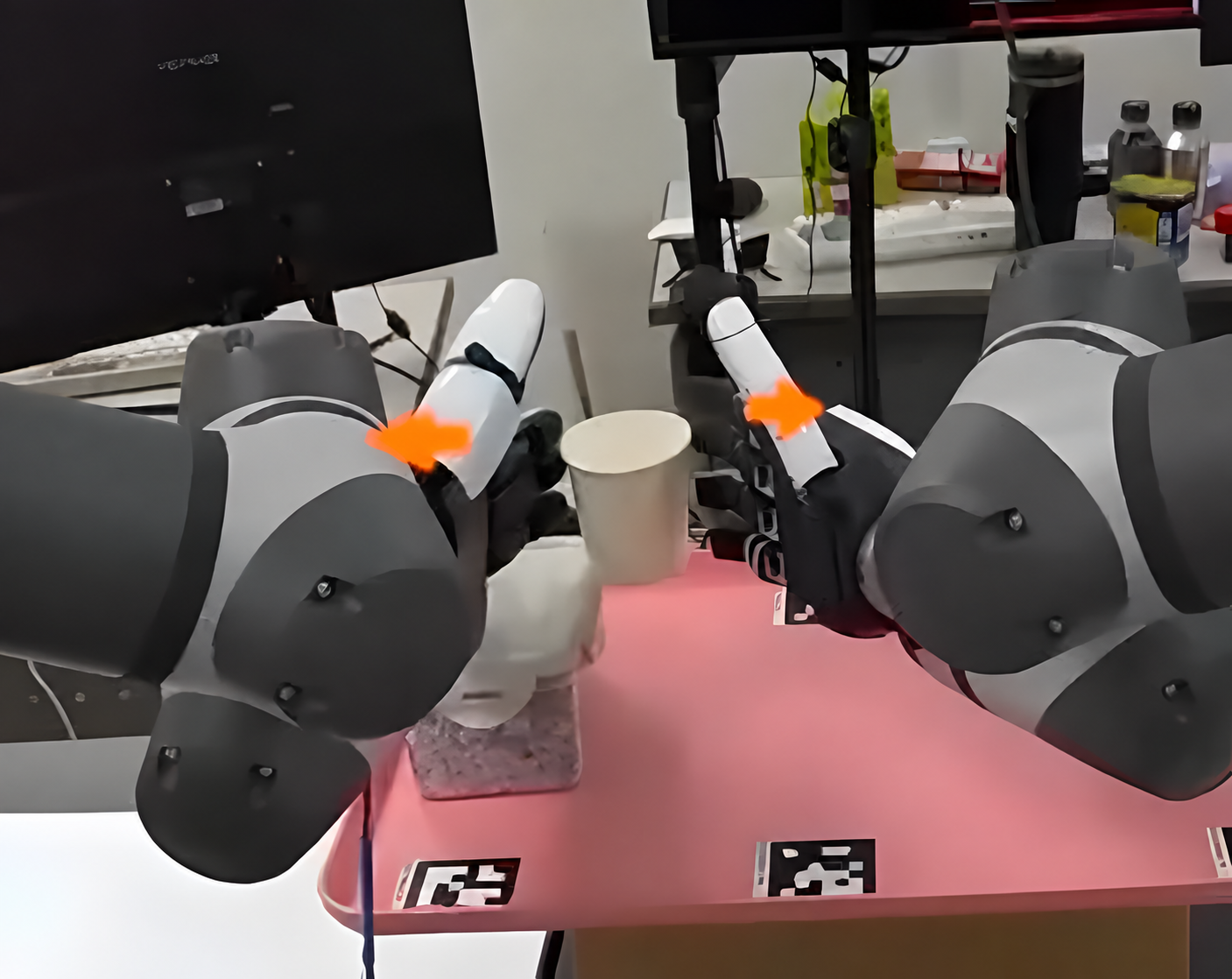}
\end{tabular}\par
\vspace{13pt}
\noindent{\fontsize{10}{12}\selectfont\bfseries (b) Task 2: Box Lifting}\par
\vspace{6pt}
\noindent\begin{tabular}{@{}m{\ExpLabelWidth}@{}c@{\hspace{\ExpGap}}c@{\hspace{\ExpGap}}c@{}}
\textbf{Baseline} &
\ExpPhoto{\ExpThreeWidth}{0.00000bp 0.00000bp 0.00000bp 126.94219bp}{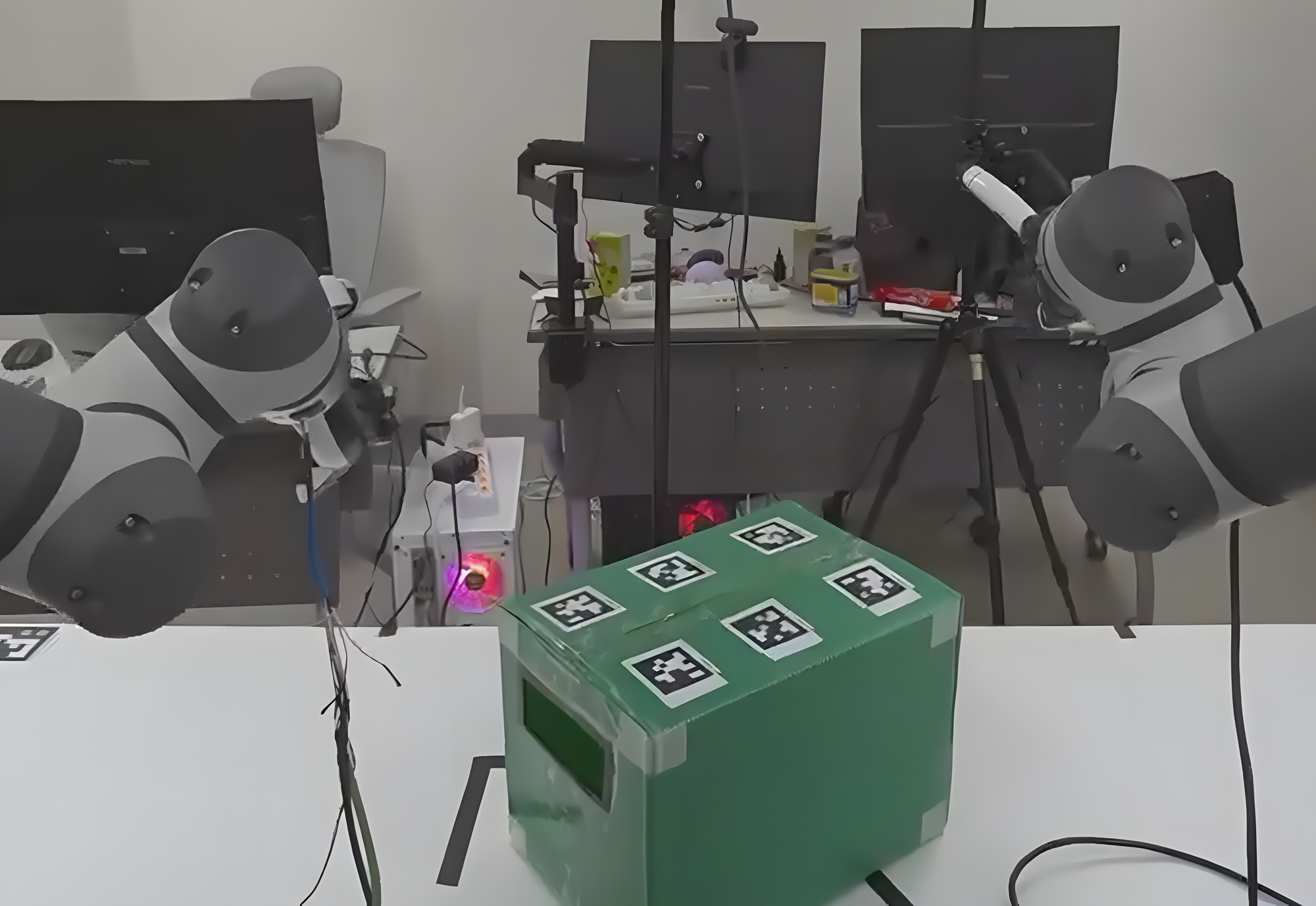} &
\ExpPhoto{\ExpThreeWidth}{0.00000bp 0.00000bp 0.00000bp 130.09653bp}{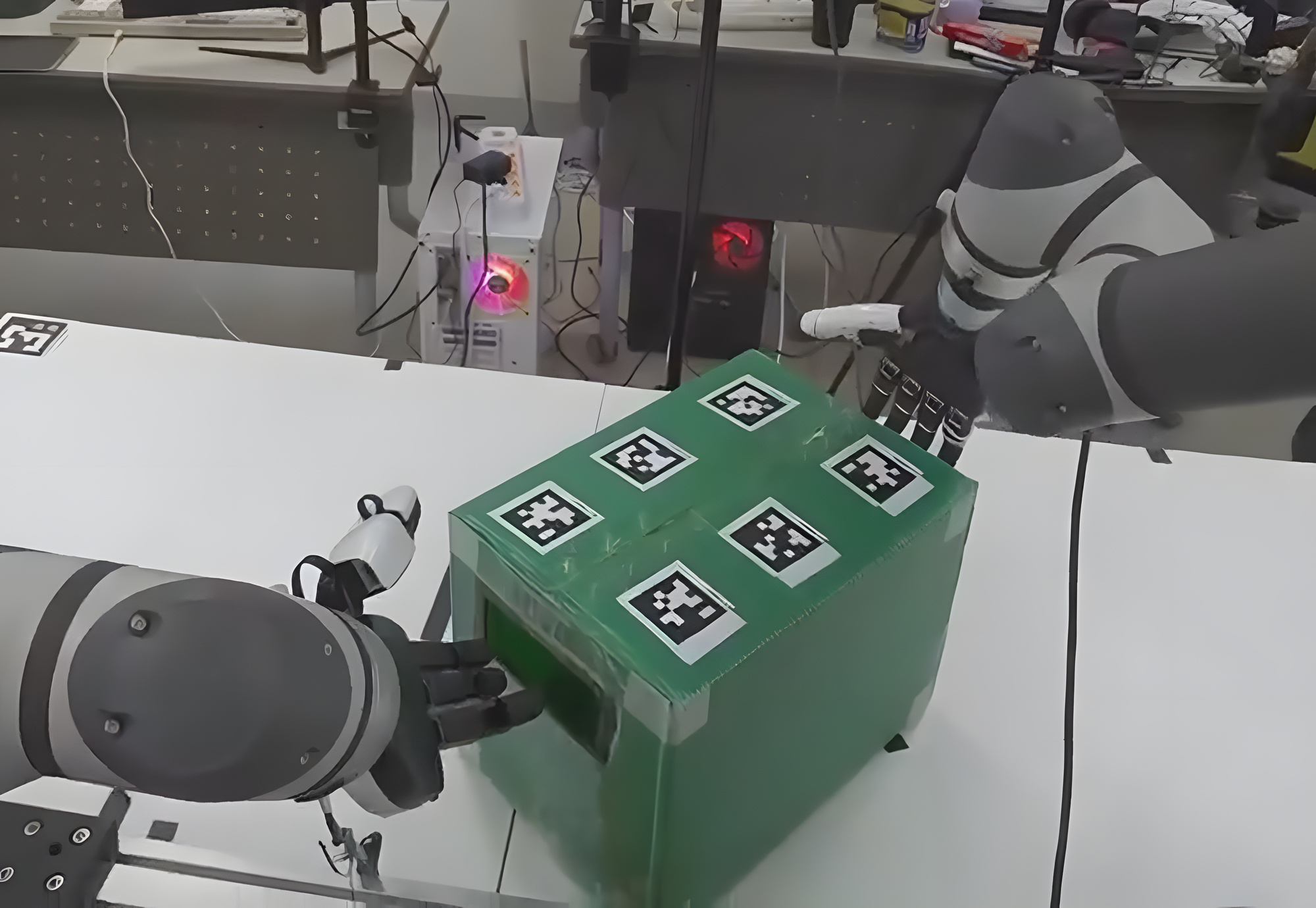} &
\ExpPhoto{\ExpThreeWidth}{0.00000bp 170.55690bp 0.00000bp 347.35690bp}{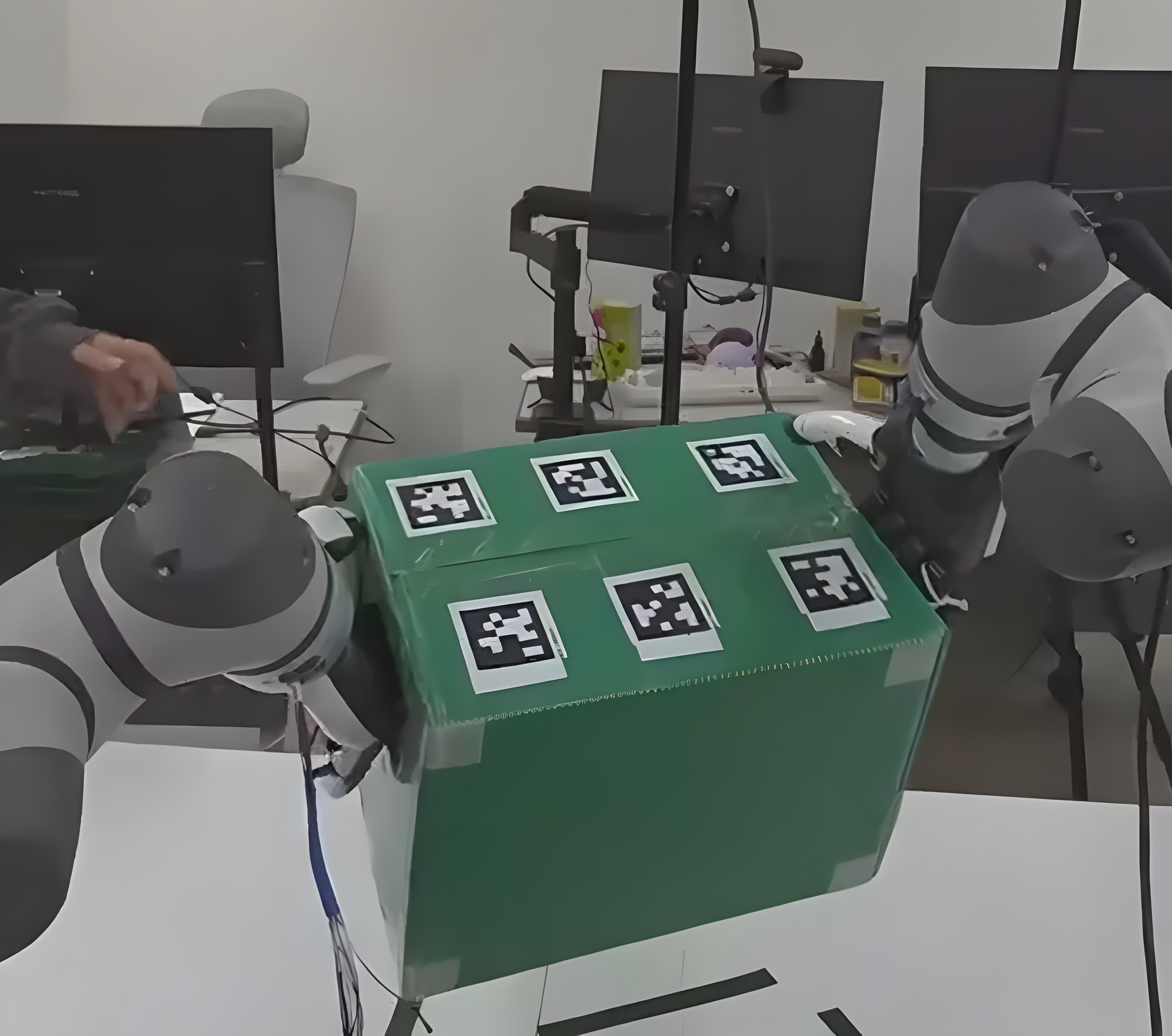} \\\noalign{\vskip 6pt}
{\fontsize{8}{10}\selectfont\bfseries User-Controlled\newline Virtual View\newline Panel} &
\ExpPhoto{\ExpThreeWidth}{100.36477bp 0.00000bp 100.36477bp 0.00000bp}{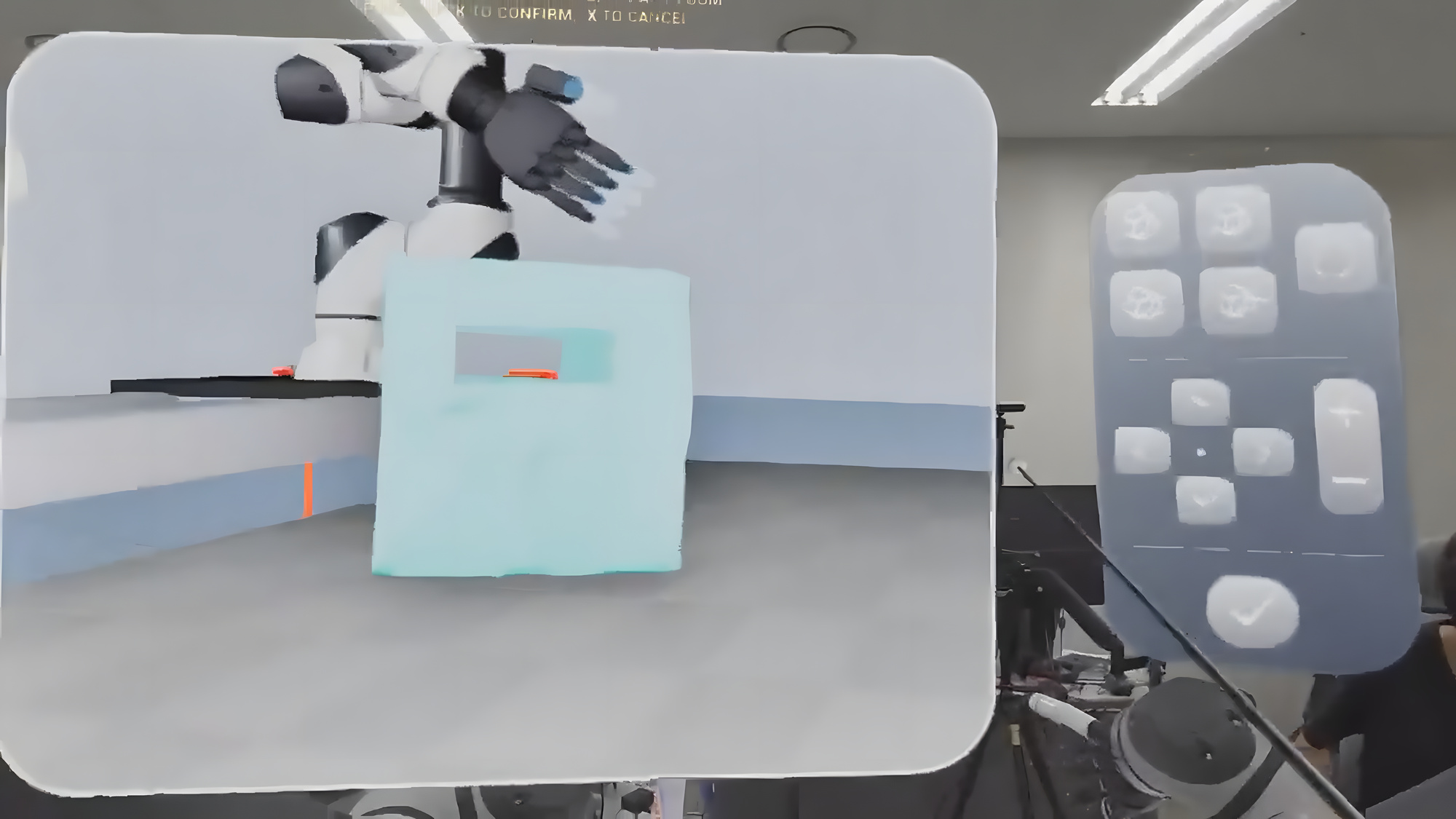} &
\ExpPhoto{\ExpThreeWidth}{0.36477bp 0.00000bp 200.36477bp 0.00000bp}{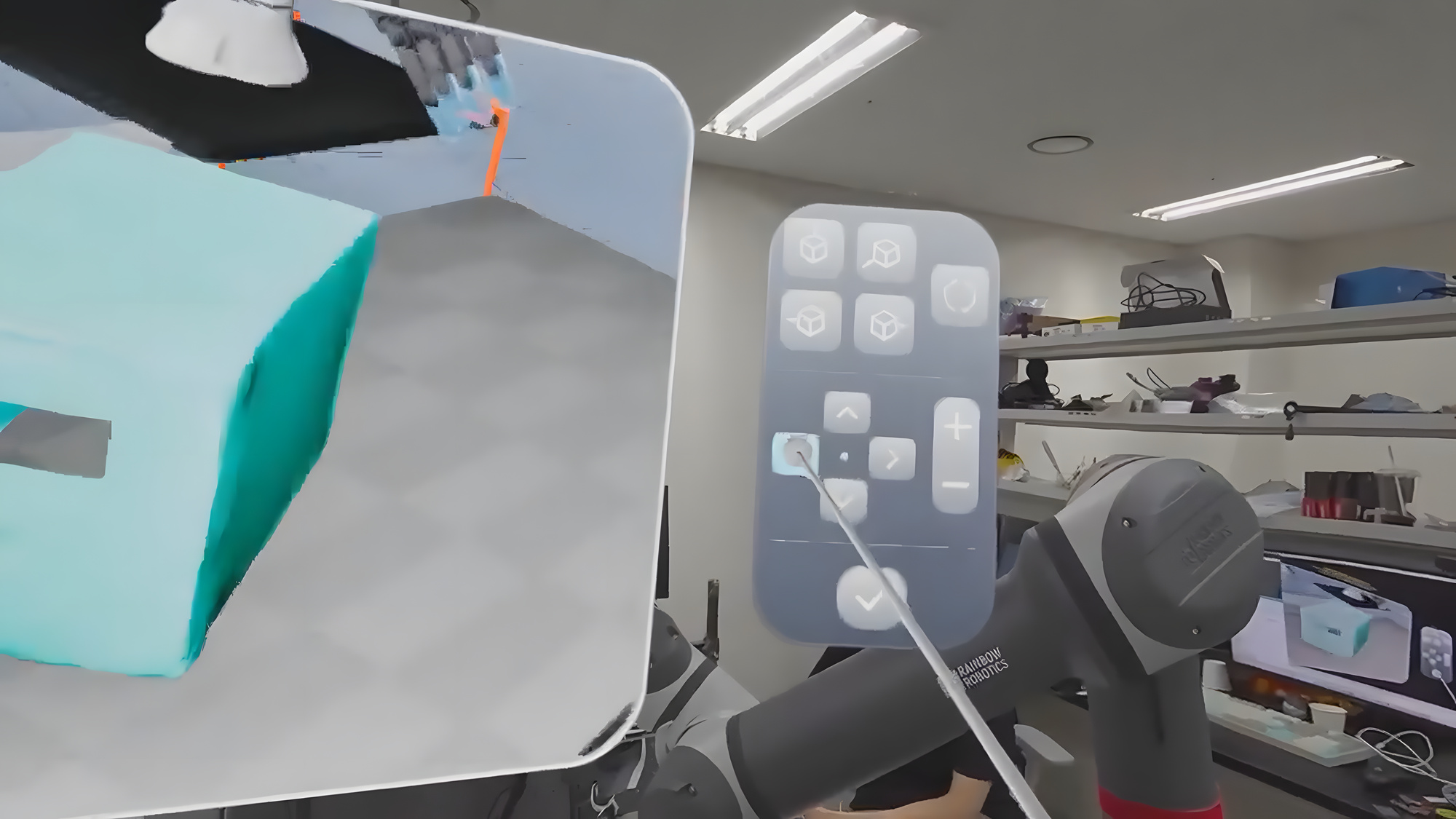} &
\ExpPhoto{\ExpThreeWidth}{100.36477bp 0.00000bp 100.36477bp 0.00000bp}{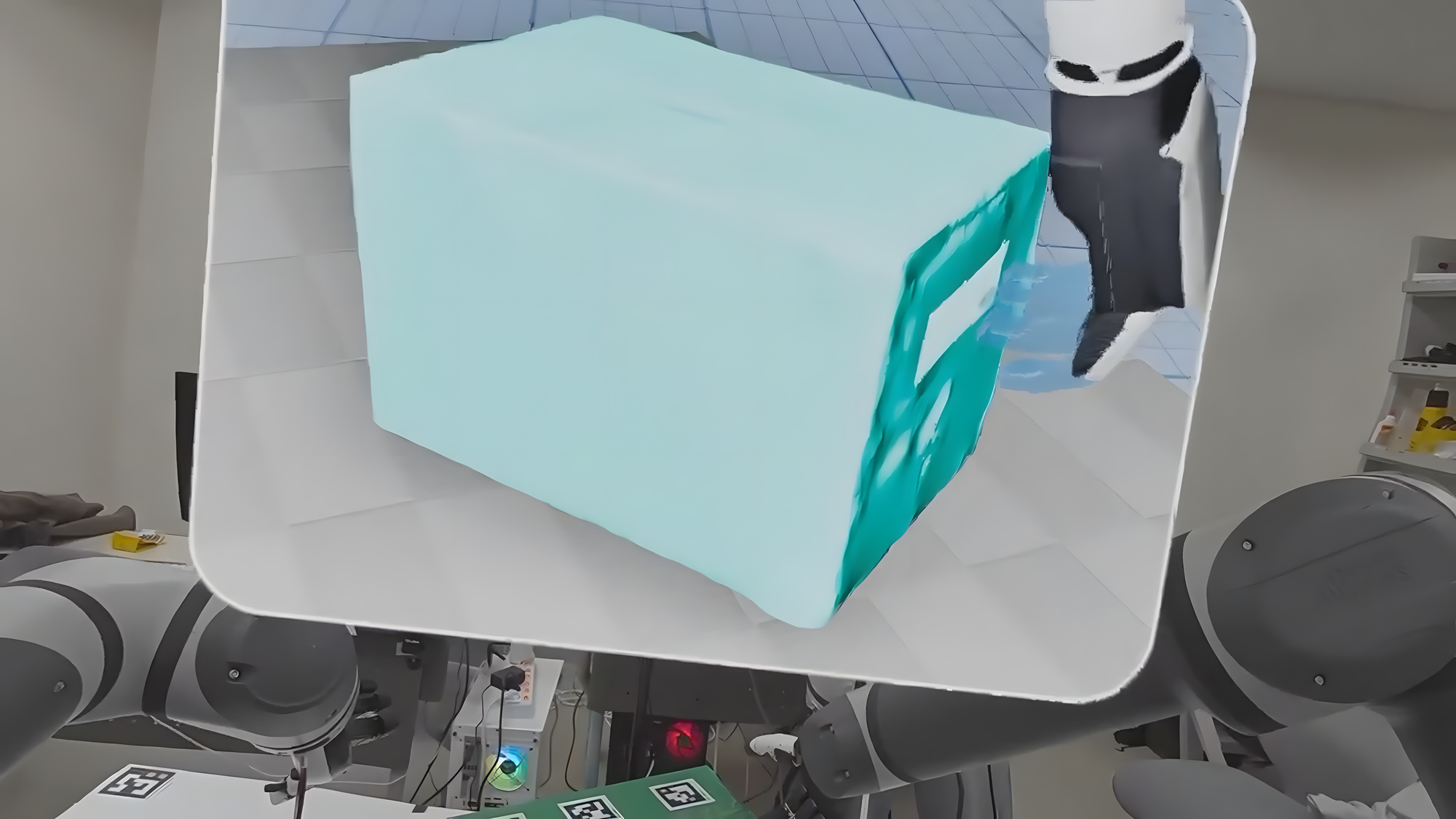} \\\noalign{\vskip 6pt}
{\fontsize{8}{10}\selectfont\bfseries Occlusion-Aware\newline Transparency\newline Rendering} &
\ExpPhoto{\ExpThreeWidth}{0.00000bp 10.50024bp 0.00000bp 91.62024bp}{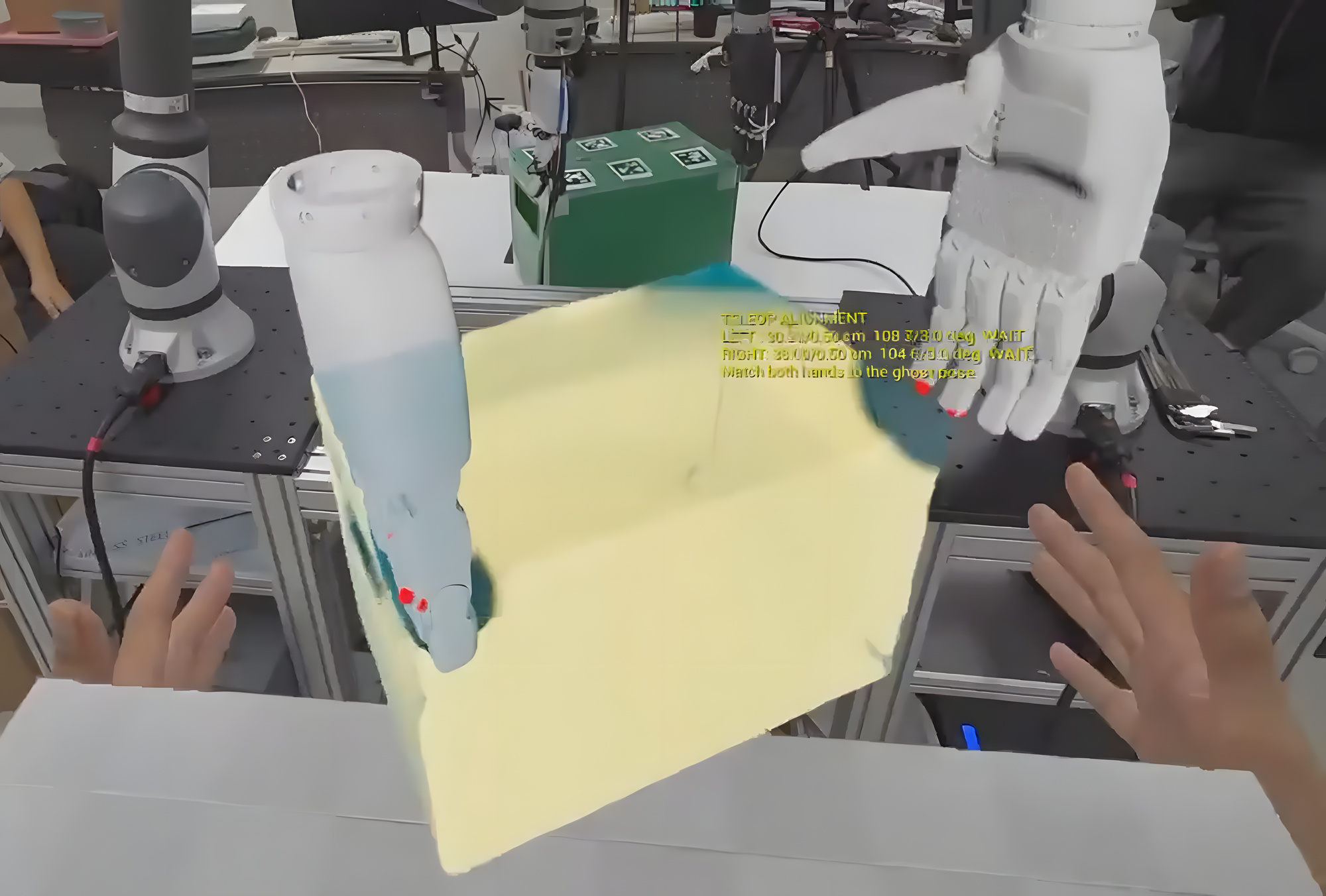} &
\ExpPhoto{\ExpThreeWidth}{0.00000bp 47.13765bp 0.00000bp 132.93765bp}{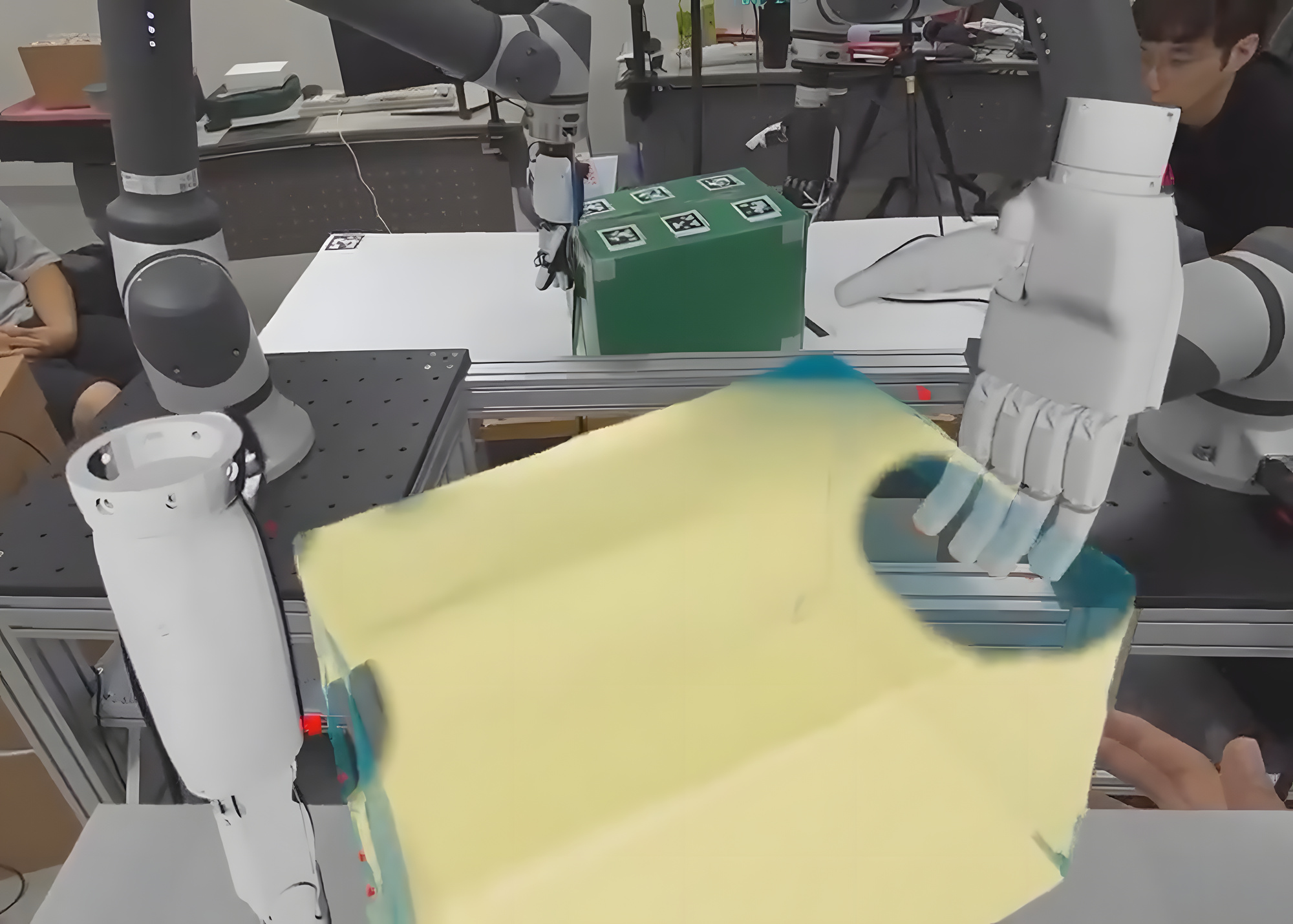} &
\ExpPhoto{\ExpThreeWidth}{100.36477bp 0.00000bp 100.36477bp 0.00000bp}{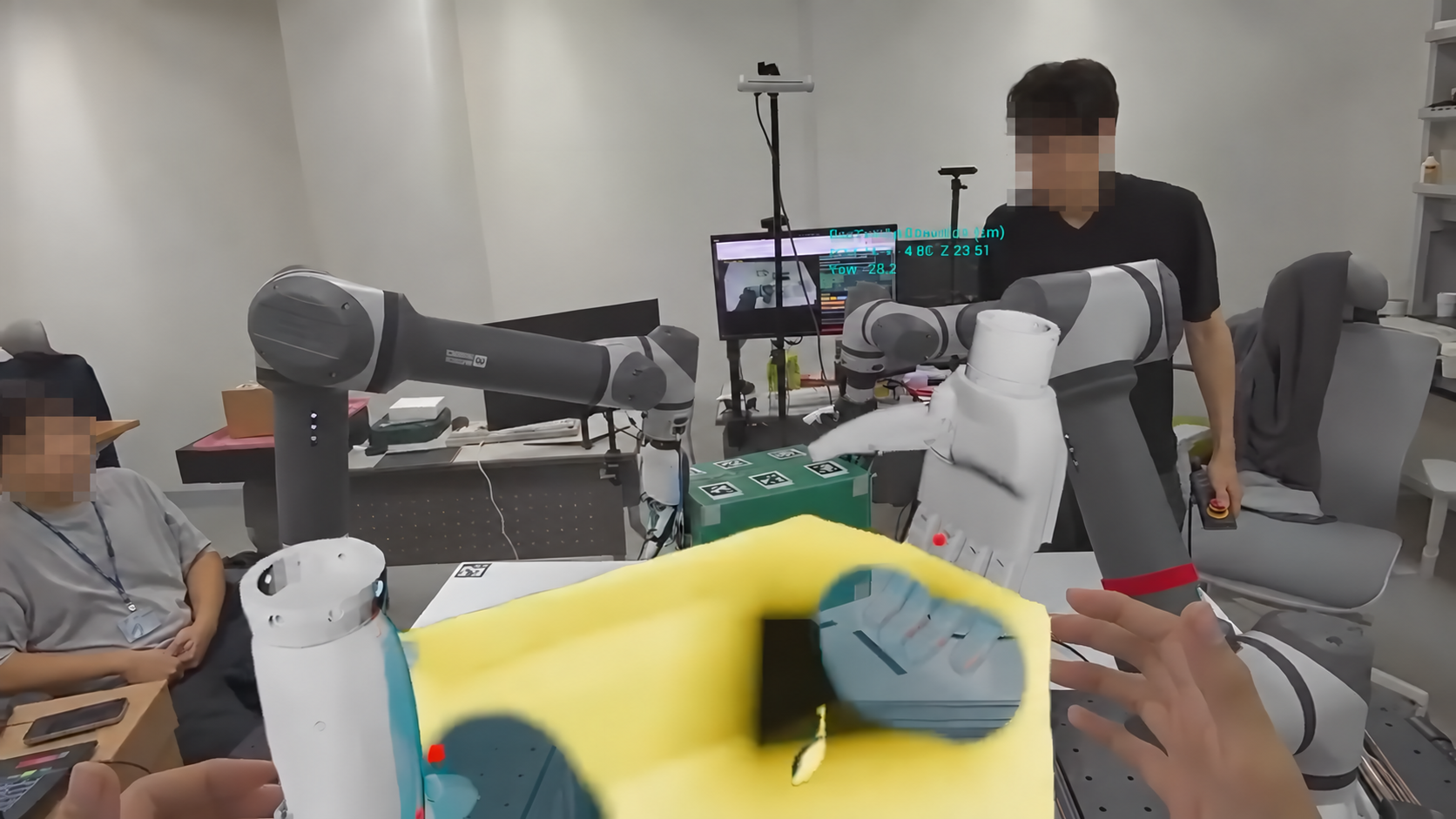}
\end{tabular}\par
\endgroup

  \caption{Experimental views across tasks and interface conditions.
  (a) Task 1 (Paper Cup Relocation): reference scene and grasping views with no visuo-force feedback,
  force heatmap feedback, and force arrow feedback.
  (b) Task 2 (Box Lifting): representative views under baseline, user-controlled virtual view panel,
  and occlusion-aware transparency rendering. Each row shows the corresponding
  experimental sequence or interface views from left to right.}
  \label{fig:experimental-interfaces}
\end{figure*}

Participants performed Task 1 and Task 2, involving precise grasp force regulation and object-induced occlusion, respectively. Figure~\ref{fig:experimental-interfaces} presents representative views of both tasks across the feedback and interface conditions.

\subsubsection{Task 1: Paper Cup Relocation}
Each participant lifted a paper cup (190 ml) with the fingertips of both hands, and transported it 450 mm to a target location, where it had to be placed upright. A paper cup is a highly deformable object; even a small applied force can cause creasing or permanent deformation. Participants were required to carefully regulate the grasp force to prevent slipping without causing creasing or permanent deformation of the cup. After each transport, the experimenter inspected the cup. A trial was considered a failure if the cup was dropped during transport, was not placed upright, or exhibited any visible crease; otherwise, it was considered a success. The cup was replaced with a new one after each failed trial.

\subsubsection{Task 2: Box Lifting}
Each participant grasped and lifted a box (\(230 \times 225 \times 320\) mm) with both hands. The box was initially rotated by 30 degrees in yaw, and this angle was maintained throughout all trials. The grasp openings of the box were located on the far side of the box, hidden from the operator's viewpoint, hindering successful grasp formation based on direct visual feedback alone. Because the fingers inserted into the openings were also occluded, participants could not visually confirm how many fingers were engaged with the box through each opening; visuo-force feedback provided this per-finger contact information, which directly determined task success. A trial was considered a success if at least three of the four non-thumb fingers of each hand were inserted into the openings, the box was lifted at least 10 cm above the ground, and it was held in that position for 4 seconds. A trial was considered a failure if two or fewer non-thumb fingers of either hand were inserted when the box was lifted, or if the box  was dropped before the 4-second hold was completed.

\section{Results}

\subsection{Task 1: Paper Cup Relocation}

\begin{figure}[!tbp]
    \centering
    \includegraphics[width=\linewidth]{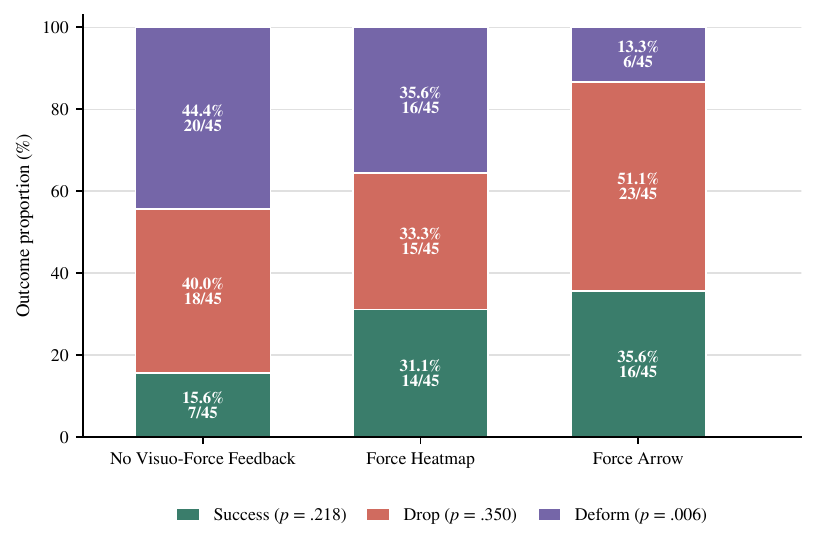}
    \caption{Distribution of task outcomes across feedback conditions. 
    The reported $p$-values compare all three conditions through
    participant-level permutation tests.}
    \label{fig:experiment1_outcomes}
\end{figure}

Across the 135 trials, force arrow feedback yields the highest observed task success rate (35.6\%, 16/45). The success rates are 31.1\% (14/45) with force heatmap feedback and 15.6\% (7/45) with no visuo-force feedback. The feedback conditions also differ in their observed failure patterns. Deformation occurs in 13.3\% of trials with force arrow feedback (6/45), compared with 35.6\% with force heatmap feedback (16/45) and 44.4\% with no visuo-force feedback (20/45). In contrast, drops are more frequent with force arrow feedback (51.1\%, 23/45) than with force heatmap feedback (33.3\%, 15/45) or no visuo-force feedback (40.0\%, 18/45). The deformation rate differs significantly across the feedback conditions ($p = .006$). However, the differences in drop rate ($p = .350$) and overall success rate ($p = .218$) are not statistically significant.

\begin{table}[!tbp]
\centering
\caption{Task 1: Effect of Visuo-Force Feedback (VFF)
(15 participants $\times$ 3 trials)}
\label{tab:exp1_force}
\scriptsize
\setlength{\tabcolsep}{4pt}
\begin{tabular}{lccc}
\toprule
 & \multirow{2}{*}{Success} & \multicolumn{2}{c}{Weighted NASA-TLX} \\
\cmidrule(lr){3-4}
VFF & & Mean & SD \\
\midrule
No Visuo-Force Feedback                 & 7/45 & 65.67 & 18.92 \\
Force Heatmap    & 14/45 & 45.78 & 23.78 \\
Force Arrow      & \textbf{16/45} & \textbf{39.67} & 25.15 \\
\bottomrule
\end{tabular}
\end{table}

Task 1 also examines whether visuo-force feedback reduces perceived workload during demonstration collection and whether this effect differs between force heatmap and force arrow representations. Table~\ref{tab:tlx_weights_task1} summarizes the NASA-TLX subscale weights for the 15 participants. A repeated-measures ANOVA of the weighted NASA-TLX scores reveals a significant effect of feedback condition ($F = 16.81$, $p < .001$; Table~\ref{tab:exp1_anova}). Bonferroni-adjusted pairwise comparisons indicate lower workload with both force heatmap ($p = .0008$) and force arrow feedback ($p = .0013$) than with no visuo-force feedback.

\begin{table}[!tbp]
\centering
\caption{Task 1: Mean Weight per NASA-TLX Subscale ($N = 15$)}
\label{tab:tlx_weights_task1}
\scriptsize
\setlength{\tabcolsep}{3pt}
\begin{tabular}{lcccccc}
\toprule
 & Mental & Physical & Temporal & Performance & Effort & Frustration \\
\midrule
Mean & 3.33 & 3.00 & 1.27 & 3.20 & 3.47 & 0.73 \\
SD   & 1.29 & 1.46 & 0.88	 & 1.74 & 0.99 & 1.16 \\
\bottomrule
\end{tabular}
\end{table}

\begin{table}[!tbp]
\centering
\scriptsize
\renewcommand{\arraystretch}{0.9}
\caption{RM-ANOVA Results: Task 1 Force Representation (1$\times$3), $N=15$}
\label{tab:exp1_anova}
\begin{tabular}{@{}llccc@{}}
\toprule
Comparison & Method & Stat & $p$ \\
\midrule
Omnibus (Mauchly $p{=}0.204$) & ANOVA & $F{=}16.81$ & $<0.0001$ \\
None vs Force Heatmap & Bonf. & $-$ & 0.0008 \\
None vs Force Arrow & Bonf. & $-$ & 0.0013 \\
Force Heatmap vs Force Arrow & Bonf. & $-$ & 0.4668 \\
\bottomrule
\end{tabular}
\par\vspace{2pt}
\end{table}

As reported in Table~\ref{tab:exp1_force}, force arrow feedback yields the lowest mean weighted NASA-TLX score (39.67), followed by force heatmap feedback (45.78) and no visuo-force feedback (65.67). However, the workload difference between force heatmap and force arrow feedback is not statistically significant ($p = .4668$). These findings support a workload reduction with either visualization relative to no visuo-force feedback, but do not establish the superiority of force arrow feedback over force heatmap feedback in terms of perceived workload.

Based on its lowest observed deformation rate in Task 1, force arrow feedback was selected for Task 2 to examine its use in combination with occlusion handling interfaces.

\subsection{Task 2: Box Lifting}

Task 2 evaluates performance in the box lifting task using success rate and insertion retries. Trial outcomes were classified according to the success and failure criteria described in the Task 2 procedure. Insertion retries were recorded by an on-site supervisor and defined as additional attempts after the initial insertion attempt; zero retries therefore indicates successful insertion on the first attempt. The analysis included 14 participants and 28 trials per condition.

Success rates were calculated over all trials in each condition. Insertion retries were summarized only for successful trials.

\begin{table}[!tbp]
\centering
\caption{Task 2: Task Performance
(14 participants; 28 trials per condition).
VFF denotes visuo-force feedback.
Insertion retries exclude the initial attempt and are averaged over
successful trials only. Bold values indicate the best performance
within each VFF condition.}
\label{tab:exp2_perf}
\scriptsize
\setlength{\tabcolsep}{4pt}
\begin{tabular}{llccc}
\toprule
 & & \multicolumn{2}{c}{Success}
 & \multirow{2}{*}{\shortstack{Mean Insertion\\Retries}} \\
\cmidrule(lr){3-4}
Interface Condition & VFF & $n/N$ & (\%) & \\
\midrule
\multirow{2}{*}{Baseline}
 & Off         & 17/28 & 60.7 & 1.47 \\
 & Force Arrow & 16/28 & 57.1 & 0.81 \\
\cmidrule(lr){1-5}
\multirow{2}{*}{\shortstack[l]{User-Controlled\\Virtual View Panel}}
 & Off         & \textbf{22/28} & \textbf{78.6} & 1.18 \\
 & Force Arrow & 20/28          & 71.4          & 1.50 \\
\cmidrule(lr){1-5}
\multirow{2}{*}{\shortstack[l]{Occlusion-Aware\\Transparency Rendering}}
 & Off         & \textbf{22/28} & \textbf{78.6} & \textbf{0.59} \\
 & Force Arrow & \textbf{23/28} & \textbf{82.1} & \textbf{0.70} \\
\bottomrule
\end{tabular}
\end{table}

Across the 168 trials, occlusion-aware transparency rendering with force arrow feedback yields the highest task success rate (82.1\%, 23/28), compared with the user-controlled virtual view panel (71.4\%, 20/28) and the baseline without occlusion handling (Baseline; 57.1\%, 16/28) under the same feedback condition. Without visuo-force feedback, occlusion-aware transparency rendering and the user-controlled virtual view panel both achieve a success rate of 78.6\% (22/28), compared with 60.7\% (17/28) for the baseline.

Occlusion-aware transparency rendering also shows the fewest insertion retries among successful trials in both feedback conditions. Without visuo-force feedback, the mean number of retries is 0.59 with occlusion-aware transparency rendering, compared with 1.18 with the user-controlled virtual view panel and 1.47 with the baseline. With force arrow feedback, the corresponding means are 0.70, 1.50, and 0.81.

The variability observed in the Baseline condition appeared to reflect individual differences in performing insertion without occlusion assistance. Some participants succeeded on the first attempt, whereas others frequently failed or required multiple retries even in successful trials.

The higher success rates of both proposed occlusion handling interfaces relative to the baseline under both feedback conditions suggest that visualizing otherwise occluded hand--object spatial relationships supports grasp formation during box lifting. In particular, occlusion-aware transparency rendering combines high success rates with the fewest insertion retries among successful trials. These descriptive results suggest that occlusion-aware transparency rendering can support manipulation with occluded grasp openings.
\FloatBarrier
\section{Conclusion}

This study examined how force feedback and spatial visualization support teleoperation in deformation-sensitive manipulation and insertion under occlusion. In Task~1, force arrow feedback yielded the lowest observed deformation rate, suggesting its potential for deformation-sensitive manipulation. However, the results did not establish a statistically significant improvement in overall task completion. The observed failure pattern suggests a possible trade-off between reducing deformation and maintaining grasp stability, motivating further work on grasp retention. Building on these findings, Task~2 examined force arrow feedback in combination with visual techniques designed to improve spatial awareness under occlusion.

In Task~2, occlusion-aware transparency rendering yielded the lowest mean number of insertion retries among successful trials, both with and without visuo-force feedback. When combined with force arrow feedback, it also achieved the highest observed success rate. Red surface markers and local object transparency provided visual references for fingertip approach locations and revealed otherwise occluded finger--opening relationships during insertion. Force arrows additionally conveyed per-finger contact information and force magnitude, allowing users to interpret these signals in relation to finger position. This combination may have supported both insertion adjustments and contact-state assessment, suggesting a potential synergy between spatial visualization and visuo-force feedback.

In the user-controlled virtual view panel condition, participants frequently selected viewpoints that did not adequately reveal the insertion state, and some reported high mental demand when using the panel. We interpret these observations as reflecting the panel's dependence on the suitability of the selected viewpoint and the additional demands that viewpoint selection and interpretation may place on some operators. Together, these findings highlight the importance of presenting spatial and force information in a shared visual context while reducing the effort required to obtain a task-relevant view during teleoperation under occlusion.

\section{Limitations}

This study evaluated the interfaces with a limited set of objects and controlled task configurations. The current implementation requires prior object reconstruction and marker-based pose tracking, which limits immediate use with unfamiliar objects or in unprepared environments. Further evaluation is therefore needed across a broader range of object geometries, occlusion conditions, and workspace configurations.

The spatial information presented by the digital twin also depends on reconstruction accuracy and the alignment and synchronization of robot and object states. Further investigation is needed to characterize how reconstruction errors, registration errors, and update latency affect users' interpretation of the rendered scene and their task performance.

\ifCLASSOPTIONcaptionsoff
  \newpage
\fi

\bibliographystyle{IEEEtran}
\bibliography{references}

\end{document}